\documentclass[10pt, conference, compsocconf]{IEEEtran}
\usepackage{amsmath, amssymb}
\usepackage[pdftex]{graphicx}
\usepackage{amsmath, amssymb}
\usepackage{blindtext}
\usepackage{hyperref}
\usepackage{listings}

\usepackage[utf8]{inputenc}
\usepackage{color, colortbl}
\usepackage{gensymb}
\usepackage{blindtext}
\usepackage{titlesec}
\usepackage{verbatim}
\graphicspath{ {figures/} }
\graphicspath{ {./images/} }

\usepackage[
backend=biber,
style=ieee,
sorting=none
]{biblatex}

\usepackage{array}
\usepackage{booktabs}
\usepackage{multirow}
\usepackage[table]{xcolor}

\usepackage[ruled, vlined]{algorithm2e}  

\ifCLASSINFOpdf
\else
\fi
\begin{document}
%
\title{Short-Term Load Forecasting Using A Particle Swarm Optimized Multi-Head Attention-Augmented CNN-LSTM Model}


\author{\IEEEauthorblockN{Paapa Kwesi Quansah\IEEEauthorrefmark{1}, Edwin Kwesi Ansah Tenkorang\IEEEauthorrefmark{2}}
\IEEEauthorblockA{Department of Electrical/Electronic Engineering\\
Kwame Nkrumah University of Science and Technology\\
Kumasi, Ghana\\}
\IEEEauthorblockA{\IEEEauthorrefmark{1}
Email: pkquansah2@st.knust.edu.gh}
\IEEEauthorblockA{\IEEEauthorrefmark{2}
Email: ekatenkorang@st.knust.edu.gh}
}


%


\maketitle

\begin{abstract}
Short-term load forecasting is of utmost importance in the efficient operation and planning of power systems, given their inherent non-linear and dynamic nature. Recent strides in deep learning have shown promise in addressing this challenge. However, these methods often grapple with hyperparameter sensitivity, opaqueness in interpretability, and high computational overhead for real-time deployment. This paper proposes an innovative approach that effectively overcomes the aforementioned problems. The approach utilizes the Particle Swarm Optimization algorithm to autonomously tune hyperparameters, a Multi-Head Attention mechanism to discern the salient features crucial for accurate forecasting, and a streamlined framework for computational efficiency. The method was subjected to rigorous evaluation using a genuine electricity demand dataset. The results underscore its superiority in terms of accuracy, robustness, and computational efficiency. Notably, its Mean Absolute Percentage Error of 1.9376 marks a significant improvement over existing state-of-the-art approaches, heralding a new era in short-term load forecasting.

\end{abstract}


%
\IEEEpeerreviewmaketitle

\section{Introduction}
In contemporary society, electrical energy has emerged as a pivotal resource propelling the economic and societal progress of nations worldwide. It is extensively utilized in industries, including manufacturing, mining, construction, and healthcare, among others. The provision of consistent and high-quality electrical power is not merely a convenience; rather, it is imperative to sustain investor confidence in economies and foster further development \cite{Frederick2014TheEO}. With the advent of new technological advancements, electricity demand has surged, creating an urgent need for more cost-effective and reliable power supply solutions \cite{rao2017energy}.

The current energy infrastructure lacks substantial energy storage capabilities in the generation, transmission, and distribution systems \cite{letcher201911}. This deficiency necessitates a precise balance between electricity generation and consumption. The maintenance of balance is contingent upon the utilization of an accurate load forecasting approach. Adapting electricity generation to dynamically meet shifting demand patterns is paramount; since failure to do so puts the stability of the entire power system at risk \cite{jiang2017hybrid}. 

Moreover, as the world pivots towards the increased adoption of renewable energy sources \cite{ellabban2014renewable}, power grids have witnessed a substantial transformation in their composition and structure. This integration of renewable energy sources, such as wind and solar power, introduces a degree of unpredictability into energy generation due to the stochastic nature of these sources \cite{ahmed2021multi}. Consequently, ensuring a stable and secure power system operation becomes an even more complex endeavor, demanding meticulous power planning and precise load forecasting.

Electric load forecasting is the practice of predicting electricity demand within a specific region. This process can be categorized into three distinct groups: short-term, medium-term, and long-term forecasting, depending on the forecasting horizon. Short-term load forecasting (STLF), which focuses on predicting electricity demand for upcoming hours, a day, or a few days, serves as the foundation for effective power system operation and analysis. It facilitates the optimization of the operating schedules of generating units, including their start and stop times, and their expected output. The accuracy of STLF is of critical importance, as it directly influences the efficient utilization of generating units \cite{chen2009short}. The absence of accurate short-term load forecasting can lead to many operational challenges, including load shedding, partial or complete outages, and voltage instability. These issues can have detrimental effects on equipment functionality and pose potential risks to human safety.

Short-term load forecasting methods are pivotal in achieving this precision. These methods can be broadly classified into two main categories: statistical methods and machine learning methods \cite{HU2021116415, KIM20191270}. Machine learning-based load forecasting methods, such as the autoregressive integrated moving average model (ARIMA) \cite{LIN2022107818}, long short-term memory (LSTM) \cite{HUANG2022108404}, generative adversarial network (GAN) \cite{HUANG2022108404}, and convolutional neural network (CNN) \cite{KIM201972}, have gained prominence. These machine learning methods excel at capturing complex nonlinear data features within load patterns \cite{LIU2020462}. They leverage the ability to discern similarities in electricity consumption across diverse power supply areas and customer types, allowing for more accurate and feasible load forecasting through the consideration of spatial-temporal coupling correlations.

\subsection{Motivation}

Based on the existing research, the following three shortcomings need to be addressed to improve the forecasting effect of the spatial–temporal distribution of the system load: (i) the lack of flexibility and scalability of traditional statistical methods, (ii) the high computational complexity of deep learning methods, and (iii) the inability of existing methods to capture the spatial-temporal correlations in load patterns.

Considering these challenges, this paper proposes a novel short-term load forecasting model that uses a particle swarm-optimized multi-head attention-augmented CNN-LSTM network. The proposed model employs a particle swarm optimization (PSO) algorithm to identify the optimal hyperparameters of the CNN-LSTM network. This enhances the model's resilience to overfitting and its accuracy. Additionally, the multi-head attention mechanism is used to learn the importance of different features for the forecasting task. Finally, a hybrid CNN-LSTM Model is used to help the system capture the spatial-temporal correlations in load patterns, hence enhancing its forecasting accuracy. 

\subsection{Contributions}

The following are the main contributions of the paper:
\begin{enumerate}
    \item \textbf{Feature Extraction:} To improve efficiency during feature extraction for STLF, PSO is employed to optimize model hyperparameters, leading to enhanced efficiency in extracting significant features with lower computational resources.
    \item \textbf{Attention-Augmented Hybrid Model:} Given that power demand is impacted by short-term fluctuations and long-term trends in data, a hybrid model is used to detect both temporal and extended dependencies, improving accuracy.
    \item \textbf{Performance evaluation:} The effectiveness of PSO-A2C-LNet has been validated using three real-world electricity demand data sets (from Panama, France, and the US). Testing results demonstrate that the PSO-A2C-LNet outperforms benchmarks in terms of forecasting performance.

\end{enumerate}

\subsection{Structure of this paper}

The rest of the paper is structured as follows. Section II provides comprehensive explanations and definitions of key terminology. Section III provides an in-depth exposition of the proposed framework and a comprehensive explanation of its operation. The findings of our tests on the framework are presented in Section IV. Section V concludes the paper.

\section{Framework Components}
\subsection{Definitions of Key Terms}
\subsubsection{Convolutional Neural Network}
A Convolutional Neural Network (CNN) is a deep learning model designed primarily for image-related tasks, but it can also be applied to other grid-like data, such as audio or time series data. CNNs are especially effective at capturing spatial dependencies within inputs. \cite{li2021survey}.

The CNN achieves the localization of spatial dependencies by using the following layers:\\

1. \textbf{Convolutional Layer}: The core operation in a CNN is the convolution operation. Convolutional layers use learnable filters or kernels to scan the input data in a localized and overlapping manner. Each filter detects specific features, like edges, textures, or more complex patterns.

Mathematically, the 2D convolution operation is defined as follows:
\begin{equation}  
(Y * X)(i, j) = \sum_{m=1}^{M}\sum_{n=1}^{N} X(i+m-1, j+n-1)\cdot Y(m, n)
\end{equation}
Here,
- \(Y\) is the filter (kernel) of size \(M \times N\).\\
- \(X\) is the input data of size \((W, H)\).\\
- \((i, j)\) represents the coordinates of the output feature map.\\
- \((m, n)\) iterates over the filter dimensions.\\

By sliding the filter across the input, the convolution operation computes feature maps that highlight different aspects of the input. This process effectively captures spatial dependencies.\\

2. \textbf{Pooling Layer}: Pooling layers are predominantly used to downsample the feature maps, reducing their spatial dimensions. Common pooling operations include max-pooling and average-pooling. Pooling aids in the invariance of network translation and minimizes the computational overhead.

For max-pooling, the operation is defined as:
\begin{equation} 
\text{Max-Pooling}(x, p, q) = \max_{i,j} x(p+i, q+j)
\end{equation}

where \(x\) is the input feature map, and \((p, q)\) represents the pooling window position. Max-pooling retains the most significant information within the window.\\

3. \textbf{Fully Connected Layer}: After multiple convolutional and pooling layers, the spatial dimensions are reduced, and the network connects to one or more fully connected layers, also known as dense layers. These layers perform classification or regression tasks by learning high-level representations.\\

Recognizing and exploiting spatial dependencies in CNNs is facilitated through several key mechanisms \cite{li2017spectral}. CNNs utilize local receptive fields, whereby each neuron is connected to a small region of the input data. This enables neurons to specialize in detecting specific features within their receptive fields, hence facilitating the network's ability to record spatial relationships across multiple scales. Additionally, weight sharing is a fundamental aspect of CNNs, where the same set of filters is applied consistently across the entire input. This weight sharing allows the network to learn translation invariant patterns, boosting its capacity to grasp spatial dependencies. Moreover, CNNs employ a hierarchical representation approach, where deeper layers in the network combine higher-level features derived from lower-level features. This hierarchical representation aids the network in comprehending complex spatial dependencies by gradually constructing abstractions. These mechanisms collectively empower CNNs to effectively model and exploit spatial dependencies in input data.\\

\subsubsection{Long Short-Term Memory Network}
The LSTM network is a type of recurrent neural network (RNN) architecture that is designed to capture and model sequential data while addressing the vanishing gradient problem that plagues traditional RNNs. LSTMs are particularly effective at locating and modeling long-term dependencies in sequential data.

LSTMs consist of multiple interconnected cells, each with its own set of gates and memory cells \cite{staudemeyer2019understanding}. The primary components of an LSTM cell are:

\textbf{Forget Gate} (\(f_t\)): Controls what information from the previous cell state (\(C_{t-1}\)) should be discarded or kept. It takes the previous cell state and the current input (\(x_t\)) as input and produces a forget gate output. 
\begin{equation}
    f_t = \sigma(W_f \cdot [h_{t-1}, x_t] + b_f)
\end{equation}

\textbf{Input Gate} (\(i_t\)): Determines what new information should be added to the cell state. It takes the previous cell state and the current input and produces an input gate output. 
\begin{equation}
    i_t = \sigma(W_i \cdot [h_{t-1}, x_t] + b_i)
\end{equation}

\textbf{Candidate Cell State} (\(\tilde{C}_t\)): This is a candidate new cell state, computed using the current input and a tanh activation function.
\begin{equation}
    \tilde{C}_t = \tanh(W_c \cdot [h_{t-1}, x_t] + b_c)
\end{equation}
\textbf{Cell State Update} (\(C_t\)): The cell state is updated by combining the information retained from the previous cell state (\(f_t \cdot C_{t-1}\)) and the new candidate cell state (\(i_t \cdot \tilde{C}_t\)).
\begin{equation}
    C_t = f_t \cdot C_{t-1} + i_t \cdot \tilde{C}_t
\end{equation}

\textbf{Output Gate} (\(o_t\)): Determines what part of the cell state should be output as the final prediction. It takes the current input and the updated cell state and produces an output gate output.
\begin{equation}
    o_t = \sigma(W_o \cdot [h_{t-1}, x_t] + b_o)
\end{equation}

\textbf{Hidden State} (\(h_t\)): The hidden state is the output of the LSTM cell, which is used as the prediction and is also passed to the next time step. It is calculated by applying the output gate to the cell state.
\begin{equation}
    h_t = o_t \cdot \tanh(C_t)
\end{equation}

LSTMs address the vanishing gradient issue of traditional RNNs by introducing key components: the cell state (\(C_t\)) and the forget gate (\(f_t\)) \cite{chandar2019towards}. The forget gate dynamically adjusts (\(f_t\)) to enable LSTMs to remember or discard information from distant time steps, facilitating the capture of long-term dependencies. Meanwhile, the cell state (\(C_t\)) acts as a memory buffer, accumulating and passing relevant information across time steps, thus enabling the model to recognize and exploit long-term patterns within input sequences.\\

\subsubsection{Multi-Head Attentional Mechanism}

The Multi-Head Attention mechanism \cite{cordonnier2020multi} is a key component of Transformer-based models, such as BERT and GPT, used for various natural language processing tasks. It excels at capturing extremely long-term dependencies in sequences of data. 

Multi-Head Attention extends the idea of the self-attention mechanism \cite{ramachandran2019stand} by employing multiple attention heads in parallel. Each attention head focuses on different parts of the input sequence, enabling the model to capture various types of information and dependencies simultaneously.

The primary components of Multi-Head Attention are as follows:

\textbf{Query (\(Q\)), Key (\(K\)), and Value (\(V\)) Projections}: For each attention head, we project the input sequence into three different spaces: query, key, and value. These projections are learned parameters.\\

\textbf{Scaled Dot-Product Attention}: Each attention head computes attention scores between the query (\(Q\)) and the keys (\(K\)) of the input sequence and then uses these scores to weight the values (\(V\)). The attention scores are computed as a scaled dot product:
\begin{equation}
\text{Attention}(Q, K, V) = \text{softmax}\left(\frac{QK^T}{\sqrt{d_k}}\right) \cdot V
\end{equation}
Here, \(d_k\) is the dimension of the key vectors.\\

\textbf{Concatenation and Linear Transformation}: After computing the attention outputs for each head, we concatenate them and apply a linear transformation to obtain the final multi-head attention output:
\begin{equation}
\text{MultiHead}(Q, K, V) = \text{Concat}(\text{head}_1, \text{head}_2, \ldots, \text{head}_h)W^O
\end{equation}
Where \(\text{Concat}\) concatenates the outputs from all attention heads, and \(W^O\) is a learned linear transformation.

\section{PSO-A2C-LNet Architecture}

\begin{figure}[!ht]
    \centering
    \includegraphics[scale=0.4]{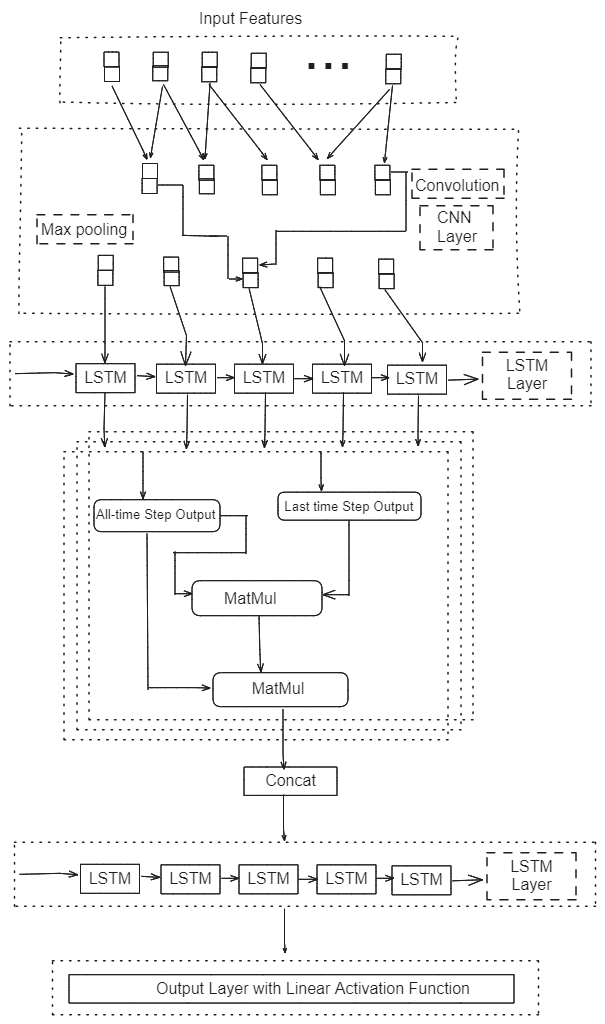}
    \caption{\textsc{PSO-A2C-LNet Structure Diagram}}
    \label{fig:figure6}
\end{figure}

PSO-A2C-LNet utilizes the various aforementioned components to extract more relevant features of the power load data to provide better predictions. This algorithm effectively improves the prediction accuracy of STLF. The model architecture is described below.\\
The model starts with an input layer designed to accept sequential data. Following the input layer, a Convolutional layer is used to capture temporal spatial patterns in the data. Subsequently, a bidirectional LSTM layer is employed to model long-term dependencies both forward and backward, enabling the capture of historical data through time. The crucial Multi-Head Attention module operates on the output of the first bidirectional LSTM layer, enabling the model to focus on the most relevant features and learn their importance. To capture intricate long-term patterns, a second bidirectional LSTM layer is employed. The final LSTM layer generates a probabilistic value. The anticipated short-term demand in kilowatts per hour is predicted by the output layer, which consists of a dense layer with one neuron and a linear activation function. A few Dropout layers were interspersed among the other model's other layers to combat overfitting. Layer Normalization is implemented subsequent to the first bidirectional layer in order to provide consistent and steady training across different inputs. The hyperbolic tangent (tanh) activation function was used for all LSTM layers.\\
To optimize model performance and convergence during training, PSO was employed to fine-tune critical hyperparameters. Table \ref{tab:optimization} shows the optimized hyperparameters and their corresponding optimization ranges.\\

\begin{table}[htbp]
  \centering
  \caption{Optimization Ranges} 
  \label{tab:optimization}
  \begin{tabular}{cc}
    \toprule
    \textbf{Hyperparameter} & \textbf{Range} \\
    \midrule
    Adaptive Learning Rate  & [0.001, 0.1] \\
    Batch Size & [1, 128] \\
    Number of Epochs & [100, 5000] \\
    Weight Initialization Techniques & (Xavier, He, Random) \\
    Loss Metrics & (MSE, Cross-Entropy, MAPE) \\
    \bottomrule
  \end{tabular}
\end{table}

The specific implementation process of the proposed algorithm is provided in Algorithm \ref{alg:example}.

\begin{algorithm}
    \caption{PSO-A2C-LNet}
    \label{alg:example}
    \KwData{Input data $D$}
    \KwResult{Output data $O$}
    
    \SetAlgoNlRelativeSize{0}

    Initialize variables\;
    Extract features $\mathbf{X}$ and target values $\mathbf{y}$ from $\mathcal{D}$\;
    Do Pre-processing\;
    Define architecture of model\;
    Start PSO\;
    \While{stopping criterion not met}{
        Find optimal parameters using PSO\; 
        Check the fitness with defined model\;
        Update variables and data structures\;
    }
    Update variables globally\;
    Train the model on the training set\;
    Evaluate the model with the three error metrics\;
    
    Post-processing steps\;
    
    \Return $O$\;
\end{algorithm}
\section{Results and Discussion}
This section comprehensively analyzes the STLF results by implementing the above model and testing it extensively on three datasets; ERCOT, RTE, and the Panama Energy Dataset. Three regression evaluation metrics are introduced for quantitative analysis of the prediction results. The performance and fitting degree of the different models are measured by the following indicators:
\begin{center}
    $R^2 = 1 - \frac{\sum_{i=1}^{n} (Y_i - \hat{Y}_i)^2}{\sum_{i=1}^{n} (Y_i - \bar{Y})^2}$
\end{center}

\begin{center}
    $\textrm{MAPE} = \frac{1}{n} \sum_{i=1}^{n} \left|\frac{Y_i - \hat{Y}_i}{Y_i}\right| \times 100\%$
\end{center}

\begin{center}
    $\textrm{MAE} = \frac{1}{n} \sum_{i=1}^{n} |Y_i - \hat{Y}_i|$
\end{center}

where $n$: Number of Observations, $Y_i$: Actual values at data point $i$, $\hat{Y}_i$: Predicted values at data point $i$ and $\bar{Y}$: Mean of the observed values.

\begin{table}[htbp]
  \centering
  \caption{Performance Metrics for Different Models on Various Datasets}
  \label{tab:performance}
  \begin{tabular}{ccccccc}
    \toprule
    \textbf{Dataset} & \textbf{Model} & $\mathbf{R^2}$ & \textbf{MAPE} (\%) & \textbf{MAE} \\
    \midrule
    \multirow{5}{*}{Panama Dataset} & A2C-LNet & 0.85 & 2.8 & 8.1 \\
    \rowcolor{blue!20}
    & \textbf{PSO-A2C-LNet} & \textbf{0.88} &\textbf{1.9} & \textbf{7.3}\\
    & Hybrid CNN-LSTM & 0.92 & 3.1 & 8.7 \\
    & Vanilla CNN & 0.81 & 3.9 & 11.2 \\
    & Vanilla LSTM & 0.86 & 3.4 & 10.0 \\
    \midrule
    \multirow{5}{*}{ERCOT Dataset} & A2C-LNet & 0.92 & 3.1 & 8.7 \\
    \rowcolor{blue!20}
    & \textbf{PSO-A2C-LNet} & \textbf{0.87} & \textbf{2.1} & \textbf{7.5} \\
    & Hybrid CNN-LSTM & 0.89 & 2.6 & 9.1 \\
    & Vanilla CNN & 0.95 & 3.2 & 7.8 \\
    & Vanilla LSTM & 0.91 & 4.0 & 8.5 \\
    \midrule
    \multirow{5}{*}{RTE Dataset} & A2C-LNet & 0.78 & 2.4 & 12.3 \\
    \rowcolor{blue!20}
    & \textbf{PSO-A2C-LNet} & \textbf{0.86} & \textbf{2.0} & \textbf{7.5} \\
    & Hybrid CNN-LSTM & 0.89 & 2.9 & 7.2 \\
    & Vanilla CNN & 0.79 & 4.2 & 12.0 \\
    & Vanilla LSTM & 0.81 & 3.1 & 11.2 \\
    \bottomrule
  \end{tabular}
\end{table}

From the results Table \ref{tab:performance}, the PSO-A2C-LNet model consistently stands out. On the Panama Energy Dataset, it achieves the highest coefficient of determination ($R^2$) at 0.88, indicating strong predictive accuracy, along with the lowest mean absolute percentage error (MAPE) of 1.9\% and the smallest mean absolute error (MAE) of 7.3, making it the top-performing model. In the ERCOT Dataset, PSO-A2C-LNet also delivers competitive results with an $R^2$ of 0.87, a MAPE of 2.1\%, and a MAE of 7.5. Similarly, on the RTE Dataset, it outperforms other models with an $R^2$ score of 0.86, a lower MAPE of 2.0\%, and a MAE of 7.5. These consistent results suggest that PSO-A2C-LNet exhibits robust predictive capabilities across diverse datasets.

While PSO-A2C-LNet excels on all datasets, the other models exhibit varying levels of performance. These comparative results emphasize the importance of model selection based on the specific dataset and application, with PSO-A2C-LNet emerging as a robust choice for diverse predictive tasks.

\begin{figure}[!ht]
    \centering
    \includegraphics[scale=0.3]{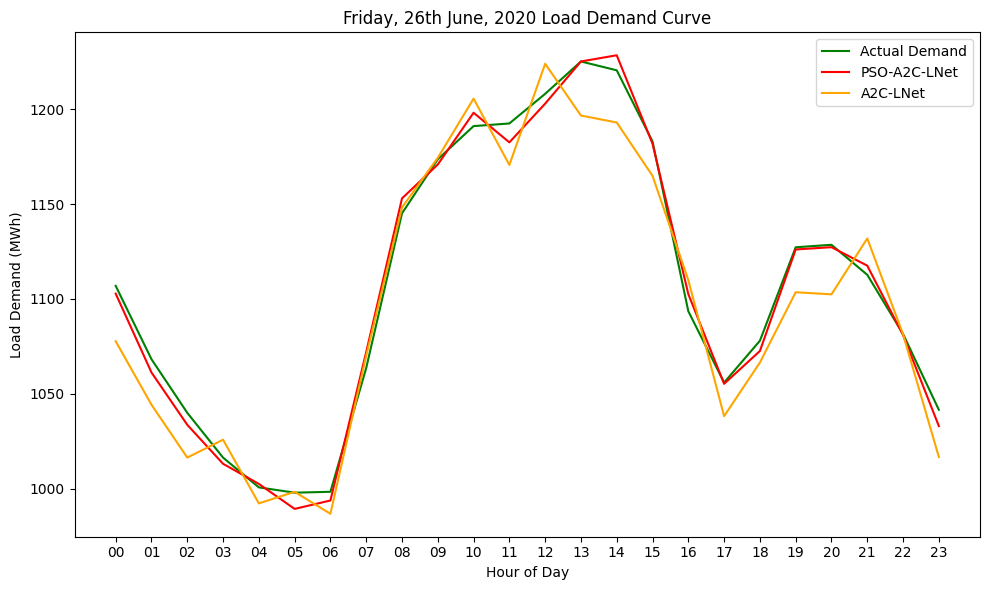}
    \caption{\textsc{Graph of Actual Demand vs. A2C-LNet vs. PSO-A2C-LNet for a Day}}
    \label{fig:figure6}
\end{figure}

\subsubsection{Comparison of results with results in literature}

Table \ref{tab:mape-results} shows the results of the A2C-LNet and the PSO-A2C-LNet on the testing dataset compared to other models in scientific literature.

\vspace{0.2in}

\begin{table}[htbp]
  \centering
  \caption{Comparison of MAPE Results with Literature}
  \label{tab:mape-results}
  \begin{tabular}{cc}
    \toprule
    \textbf{Model} & \textbf{Best MAPE Results} \\
    \midrule
    A2C-LNet  & 2.8097 \\
    \rowcolor{blue!20}

    \textbf{PSO-A2C-LNet (Our Proposed Model)} & \textbf{1.9376} \\
    Integrated CNN and LSTM Network \cite{9358156} & 3.49 \\

    LSTM network considering attention mechanism\cite{LIN2022107818} & 2.26 \\

    ANN-IEAMCGM-R \cite{SINGH2019460} & 3.59 \\

    TCN-LightGBM \cite{9210509} & 2.64 \\

    nonAda-GWN \cite{9468666} & 7.42 \\

    Ada-GWN \cite{9468666} & 6.83 \\

    Stacked XGB-LGBM-MLP \cite{9468666} & 2.69 \\

    GRU-CNN Hybrid Neural Network Model \cite{Wu2020-wt} & 2.8839 \\
    \bottomrule
  \end{tabular}
\end{table}

\section{Conclusion}
In conclusion, this research paper has introduced a novel neural network architecture for short-term load forecasting, amalgamating Convolutional Neural Network and Long Short-Term Memory models, reinforced by a Multi-Head Attention Mechanism. Empirical assessments confirm its superiority over traditional methods and standalone neural network models, with demonstrated applicability to real-world datasets. 

Future work will focus on optimizing the proposed architecture, exploring further hyperparameter tuning, and investigating additional data preprocessing techniques for enhanced forecasting. Additionally, integrating robust data privacy measures, such as federated learning or secure enclaves, into the architecture is essential to address emerging privacy concerns in load forecasting, ensuring secure and privacy-preserving predictions while advancing the scalability and adaptability of the framework to diverse forecasting challenges and datasets.


\section*{Decalaration of Competing Interest}
The authors declare that there is no conflict of interest regarding the publication of this paper.

\section*{Acknowledgment}
The authors would like to thank Professor Philip Yaw Okyere for guiding the research.



%

\printbibliography

\end{document}